\documentclass[letterpaper]{article}
\usepackage{uai2020}
\usepackage[margin=1in]{geometry}

\usepackage{times}

\usepackage{hyperref}       
\usepackage{url}            
\usepackage{booktabs}       
\usepackage{amsfonts}       
\usepackage{nicefrac}       
\usepackage{microtype}      
\usepackage{graphicx}
\usepackage{subcaption}
\usepackage{amsmath}
\usepackage{amssymb}
\usepackage{tabularx}
\usepackage[ruled, vlined,linesnumbered]{algorithm2e}

\usepackage{cleveref}

\crefformat{section}{\S#2#1#3} 
\crefformat{subsection}{\S#2#1#3}
\crefformat{subsubsection}{\S#2#1#3}

\usepackage[citestyle=authoryear-comp, bibstyle=authoryear, maxcitenames=2, maxbibnames=8, natbib=true, backend=bibtex, giveninits=true, hyperref]{biblatex}
\defbibheading{bibliography}[References]{\subsubsection*{#1}}
\bibliography{ref}

\usepackage[dvipsnames]{xcolor}
\usepackage{tikz}
\usetikzlibrary{bayesnet}
\usetikzlibrary{shapes.geometric, arrows, shadows}
\newcommand{\ra}[1]{\renewcommand{\arraystretch}{#1}}
\usepackage{bm}
\usepackage{nccmath}

\makeatletter
\newcommand*{\indep}{%
  \mathbin{%
    \mathpalette{\@indep}{}%
  }%
}
\newcommand*{\nindep}{%
  \mathbin{
    \mathpalette{\@indep}{\not}
  }%
}
\newcommand*{\@indep}[2]{%
  \sbox0{$#1\perp\m@th$}
  \sbox2{$#1=$}
  \sbox4{$#1\vcenter{}$}
  \rlap{\copy0}
  \dimen@=\dimexpr\ht2-\ht4-.2pt\relax
  \kern\dimen@
  {#2}%
  \kern\dimen@
  \copy0 
} 
\makeatother

\makeatletter 
\g@addto@macro{\@algocf@init}{\SetKwInOut{Parameter}{Parameters}} 
\makeatother

\DeclareMathOperator*{\argmax}{arg\,max}
\DeclareMathOperator*{\argmin}{arg\,min}

\title{Semi-supervised learning, causality, and the conditional cluster assumption}

\author{
Julius von K\"ugelgen$^{1,2}$ \quad\quad\quad\quad
Alexander Mey$^3$ \quad\quad\quad\quad
Marco Loog$^{3,4}$ \quad\quad\quad\quad
Bernhard Sch\"olkopf $^1$\\[0.75em]
$^1$ Max Planck Institute for Intelligent Systems T\"ubingen, Germany\\
$^2$ Department of Engineering, University of Cambridge, United Kingdom\\
$^3$ Delft University of Technology, The Netherlands\\
$^4$ University of Copenhagen, Denmark\\
\texttt{\{jvk, bs\}@tuebingen.mpg.de}, \texttt{\{a.mey, m.loog\}@tudelft.nl}
}

\newcommand{\negspace}{-0.5em}
\newcommand{\negspacefig}{-0.3em}
\newcommand{\negspacesub}{-0.3em}
\newcommand{\negspacesubsub}{-0.25em}
\newcommand{\negspacepar}{-0.em}

\begin{document}

\maketitle

\begin{abstract}
While the success of semi-supervised learning (SSL) is still not fully understood, Sch\"olkopf et al.\ (2012) have established a link to the principle of independent causal mechanisms. They conclude that SSL should be impossible when predicting a target variable from its causes, but possible when predicting it from its effects.
Since both these cases are restrictive, we extend their work by considering classification using cause and effect features at the same time, such as predicting a disease from both risk factors and symptoms. While standard SSL exploits information contained in the marginal distribution of all inputs (to improve the estimate of the conditional distribution of the target given inputs), we argue that in our more general setting we should use information in the conditional distribution of effect features given causal features. We explore how this insight generalises the previous understanding, and how it relates to and can be exploited algorithmically for SSL.
\end{abstract}

\vspace{\negspace}
\section{INTRODUCTION}
\label{sec:introduction}
\vspace{\negspace}
Due to the scarcity and often high acquisition cost of labelled data, machine learning methods that make effective use of large quantities of unlabelled data are crucial.
One such method is semi-supervised learning (SSL) \citep{Zhu2005,Chapelle2010} where, in addition to labelled data, possibly large numbers of unlabelled observations are available to the learner at training time.
While positive results have been obtained on a range of problems, a shortcoming is that SSL can actually degrade performance if certain assumptions are not met \citep[]{Chapelle2010}.
For example, \citet[]{BenDavid2008} show that the cluster assumption, commonly used in SSL settings, can lead to degraded performance even in simple cases, e.g., for binary classification with data generated from two unimodal Gaussians.
Such examples make it clear that many aspects of SSL are, as of yet, not well understood. 

Building on the \emph{principle of independent causal mechanisms} (ICM) \citep{daniuvsis2010inferring, Peters2017}, \citet{Schoelkopf2012} have pointed out a link between the possibility of SSL and the causal structure underlying a given learning problem.
Specifically, they argue that SSL should be impossible when predicting a target variable from its causes (\emph{causal learning}), but possible when predicting it from its effects (\emph{anticausal learning})---see \cref{sec:background} for details. Empirical evidence from a meta-analysis of various SSL scenarios supports this claim.

In this work, we extend the investigation of connections between SSL and causality to a more general setting.
Rather than treating causal and anticausal learning in isolation, we consider {\em predicting a target variable from both causes and effects  at the same time}.
As an example, consider the setting of predicting disease from medical data where both types of features are commonly found:
a patient's age, sex, medical family history, genetic information, diet, and other risk factors such as smoking all constitute (possible) causal features; examples of effect features, on the other hand, include the clinical symptoms exhibited by the patient, as well as results of medical tests such as imaging results, serum tests, or tissue samples.

As our main result, we show in \cref{sec:method} that for this setting of SSL with both cause and effect features, \emph{the relevant information that additional unlabelled data may provide for prediction is contained in the conditional distribution of effect features given causal features}.
This generalisation of \citet[]{Schoelkopf2012} contains their results for causal and anticausal learning as special cases. 
We then use this new insight to reformulate classical SSL assumptions in \cref{sec:assumptions}, and propose algorithms based on these assumptions in \cref{sec:algorithms}. 
Results from evaluating our methods against well-established SSL algorithms on synthetic and medical datasets in \cref{sec:experiments} empirically support our analysis.
We critically discuss our assumptions and results in~\cref{sec:discussion} and conclude with an outlook on future work in~\cref{sec:conclusion}.

\section{BACKGROUND \& RELATED WORK}
\label{sec:background}
\vspace{\negspace}
Throughout, we use $X$ to denote a random variable taking values in $\mathcal{X}\subseteq\mathbb{R}^d$. $P$ denotes a probability measure and $P(X)$ the probability distribution of $X$ with density $p$. We write $x\in \mathbb{R}$ for a scalar, $\mathbf{x}\in \mathbb{R}^d$ for a vector, and $\mathbf{X}\in \mathbb{R}^{n\times d}$ for a matrix or collection of samples.

\subsection{SEMI-SUPERVISED LEARNING (SSL)} \label{sec:background_ssl}
\vspace{\negspacesub}
SSL describes a learning setting where, in addition to a labelled sample $(\mathbf{X}^l, \mathbf{y}^l)=\{(\mathbf{x}^i, y^i)\}_{i=1}^{n_l}$, we have access to an unlabelled sample $\mathbf{X}^u=\{\mathbf{x}^i\}_{i=n_l+1}^{n_l+n_u}$ from the same distribution $P$ at training time.\footnote{In the statistics literature this setting is also known as data \textit{missing completely at random} (MCAR).}
It is usually assumed that $n_l\ll n_u$.
At test time, the task is to predict targets $Y$ from inputs $X$.
If predictions are made on the unlabelled training data only we speak of \emph{transductive} learning \citep{vapnik1998}. 
The aim and hope of SSL is that additional unlabelled data helps in making better predictions.
$\mathbf{X}^u$ can improve the estimate of $P(X)$, but SSL aims at improving  $P(Y|X)$.  This can only work if there is a link between $P(X)$ and $P(Y|X)$. Indeed, many approaches to SSL establish such a link through additional assumptions \citep{Zhu2005, Chapelle2010, mey2019improvability}.
Two common ones are the \emph{cluster assumption}, positing that points in the same cluster of $P(X)$ have the same label $Y$;
and \emph{low-density separation}, stating that class boundaries of $P(Y|X)$ should lie in an area where $P(X)$ is small. For original references, as well as for a discussion of how these assumptions relate to various SSL methods refer to \citet{Chapelle2010}.

We briefly mention four of the more common methods, starting with self-learning (sometimes also called the Yarowsky-algorithm).  This is a wrapper algorithm that initialises the learner based on the labelled data, updates the labels for the unlabelled data, and then retrains based on all labelled data available, possibly iterating this procedure \citep{Scudder65, Blum1998, Abney2004}.  Secondly, generative model approaches maximise the likelihood of a generative model
\begin{align}
\begin{split}
\label{eq:generative_modelSSL}
\textstyle \hat{\bm{\theta}}^{\text{MLE}}
&=\argmax_{\bm{\theta}}p(\mathbf{X}^l,\mathbf{y^l},\mathbf{X}^u|\bm{\theta})\\
&= \argmax_{\bm{\theta}} \sum_{\mathbf{y}^u\in\mathcal{Y}^{n_u}}p(\mathbf{X}^l,\mathbf{y^l},\mathbf{X}^u, \mathbf{y}^u|\bm{\theta}).
\end{split}
\end{align}
While this is a hard optimisation problem due to the latent variables $\mathbf{y}^u$, a local optimum can be found via the expectation maximisation algorithm (EM) \citep{dempster1977maximum}.  The third class of common methods are the graph-based approaches. These construct a similarity-based graph representation of the data and propagate labels to neighbours in this graph \citep{zhu2002learning, zhu2003semi, zhou2004learning}.  Finally, transductive SVMs maximise a (soft) margin over labelled and unlabelled data while minimising a regularised risk on the labelled data~\citep{vapnik1998,joachims1999transductive}.

\subsection{CAUSALITY}\label{sec:background_causality}
\vspace{\negspacesub}
Despite data showing a positive correlation between chocolate consumption and the number of Nobel prizes per capita \citep{messerli2012chocolate}, we would not expect that force-feeding the population with chocolate would result in higher research output.
The correlation in this example may make chocolate consumption a useful predictor in an i.i.d.\ setting, but it does not allow one to answer interventional questions of the form ``\textit{what would happen if we actively changed some of the variables?}''.

This notion of intervention is at the heart of the difference between correlation and causation.
While much of machine learning is concerned with using correlations between variables to make predictions, \citet{reichenbach1956direction} has argued that such correlations always result from underlying causal relationships: statistical dependence is an epiphenomenon---a by-product of a causal process.\footnote{For the given example, a possible explanation for the observed correlation is a healthy economy acting as common cause for both chocolate consumption and a good education system.}

\paragraph{Structural causal model (SCM)}
To reason about causality in SSL, we adopt the structural causal model (SCM) framework \citep{Pearl2000} which defines a causal model over a set of observed variables $\{X_1, ..., X_d\}$ to consist of (i) a collection of structural assignments,
\begin{equation}
\label{eq:SCM_definition}
X_i := f_i(\text{PA}_{i}, N_i) \quad\quad \text{for} \quad\quad i=1,...,d,
\end{equation}
where $f_i$ are deterministic functions computing $X_i$ from its causal parents $\text{PA}_i\subseteq \{X_1,...,X_d\}\setminus X_i$ and an exogenous noise variable $N_i$; and (ii) a factorising joint distribution $P$ over the unobserved noise variables,
\begin{equation*}
P(N_1, ..., N_d)=\prod_{i=1}^d P(N_i).
\end{equation*}
This assumption of mutually independent noises entails that all causes common to any pair of observed variables are included in the model (i.e., there are no hidden confounders), and is referred to as \textit{causal sufficiency}.
Together, (i) and (ii) define a causal generative process and imply an observational joint distribution over $X_1, ..., X_d$ which factorises over the induced causal graph\footnote{The induced causal graph $\mathcal{G}$ is obtained by drawing a directed edge from each node in $\text{PA}_i$ (i.e, the direct causes of $X_i$) to $X_i$ for all $i$. We assume throughout that $\mathcal{G}$ is acyclic.} $\mathcal{G}$ as:
\begin{equation}
\label{eq:causalmarkov}
P(X_1,\dots,X_d)=\prod_{i=1}^d P(X_i|\text{PA}_{i}).
\end{equation}

\paragraph{Principle of independent causal mechanisms (ICM)}

\begin{figure}[]
    \centering
    \includegraphics[width=\columnwidth]{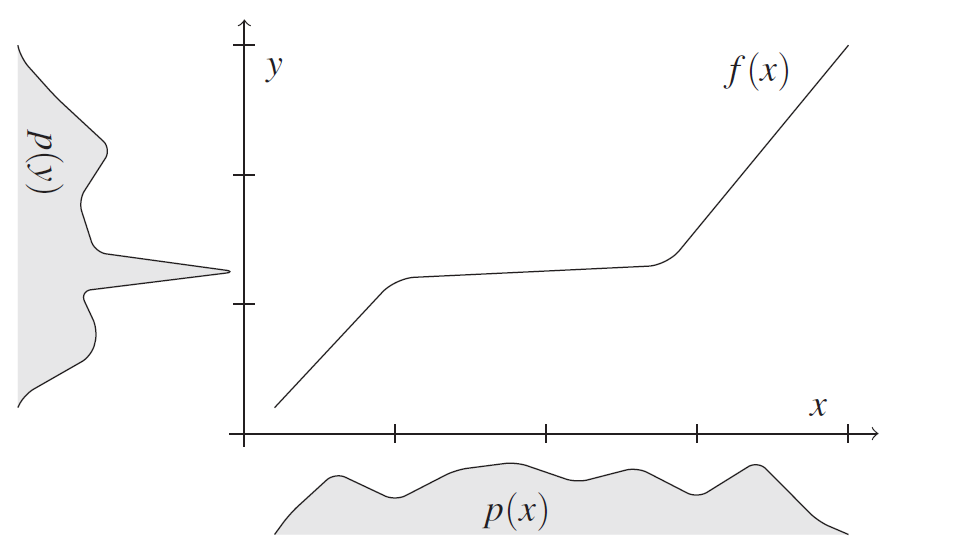}
    \caption{Illustration of the ICM principle for the setting $X\rightarrow Y$. If the distribution of the cause $p(x)$ is chosen independently of the mechanism $f:\mathcal{X}\rightarrow\mathcal{Y}$ representing $p(y|x)$, then such independence is violated in the backward (non-causal) direction: $p(y)$ has large density where $f$ has small slope and thereby contains information about $f^{-1}=p(x|y)$.
    Figure from~\citet{janzing2012information}.}
    \label{fig:IGCI}
    \vspace{\negspacefig}
\end{figure}

Motivated by viewing the $f_i$ in~\eqref{eq:SCM_definition} as independent physical mechanisms of nature, the principle of independent causal mechanisms (ICM) states that ``\emph{the causal generative process (...) is composed of independent and autonomous modules that do not inform or influence each other}'' \citep{Peters2017}.
In other words, the conditional distributions of each variable given its causal parents, $P(X_i|\text{PA}_i)$ in \eqref{eq:causalmarkov}, are independent objects which do not share any information. 
Importantly, this notion of independence is different from \textit{statistical} independence of random variables (indeed, the variables can still be statistically dependent).
Instead, it should be understood as an \emph{algorithmic} independence on the level of distributions.

Intuitively, two distributions are algorithmically independent if encoding them jointly does not admit a shorter description than describing each of them separately.
In this case, we say that they do not share information.
This notion has been formalised in terms of Kolmogorov complexity (or algorithmic information) $K(\cdot)$ by \citet{Janzing2010}: two distributions $P(X)$ and $P(Y|X)$ are considered algorithmically independent if and only if\ 
\begin{equation}
\label{eq:kolmogorov}
K\big(P(X,Y)\big) \overset{+}{=}  K\big(P(X)\big) + K\big(P(Y|X)\big),
\end{equation}
where the notation $\overset{+}{=}$ refers to a constant due to the choice of a Turing machine in the definition of algorithmic information.
Note that the RHS of \eqref{eq:kolmogorov} is always greater or equal to the LHS, and equality indicates that there is no redundant information which could be compressed by describing the two distributions jointly.

Crucially, if the ICM principle holds, i.e., the conditionals in the causal factorisation \eqref{eq:causalmarkov} do not share information, this independence is generally violated for other non-causal factorisations.
This is illustrated in Fig.~\ref{fig:IGCI} for the bivariate setting\footnote{where ICM reduces to an \emph{independence of cause and mechanism} \citep{daniuvsis2010inferring,Lemeire2006}} of  $X\rightarrow Y$\textbf{}
 using the IGCI model of \citet{janzing2012information} in which a deterministic, invertible function $f$ generates effect $Y$ from cause $X$.
If the input distribution of the cause, $p(x)$, is chosen independently from the mechanism $f(x)$ (or more generally, $p(y|x)$), then this independence is violated in the backward (non-causal) direction, since $p(y)$ has a large density where $f$ has small slope (see Fig.~\ref{fig:IGCI}), and $p(y)$ thereby contains information about $f^{-1}(y)=p(x|y)$.\footnote{The amount of shared information can be quantified as $\text{cov}[\log g', p_Y]:=\int_0^1 \log g'(y)p_Y(y) dy - \int_0^1 \log g'(y)dy \geq 0$, where $g=f^{-1}$, with equality iff.\ $f$ is the identity \citep{janzing2012information}; note that, by assumption, $\text{cov}[\log f', p_X]=0$.
}

\subsection{SSL IN CAUSAL AND ANTICAUSAL LEARNING SETTINGS} \label{sec:causalanticausal}
\vspace{\negspacesub}

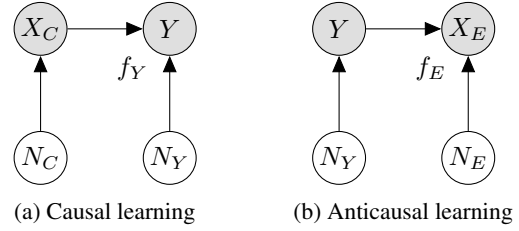
\begin{figure}[]
    \begin{subfigure}[b]{.5\columnwidth}
        \centering
        \begin{tikzpicture}
            \centering
            \node (X_C) [obs] {$X_C$};
            \node (Y) [obs, right=of X_C] {$Y$};
            \node (f_Y) [const, right=of X_C, yshift=-1.5em,xshift=-1em] {$f_Y$};
            \node (N_C) [latent, below=of X_C] {$N_C$};
            \node (N_Y) [latent, below=of Y] {$N_Y$};
            \edge[] {X_C, N_Y} {Y};
            \edge {N_C} {X_C};
        \end{tikzpicture}
        \caption{Causal learning}
        \label{fig:causal_learning}
    \end{subfigure}%
    \begin{subfigure}[b]{0.5\columnwidth}
        \centering
        \begin{tikzpicture}
            \centering
            \node (Y) [obs] {$Y$};
            \node (X_E) [obs, right=of Y] {$X_E$};
            \node (f_E) [const, right=of Y, yshift=-1.5em,xshift=-1em] {$f_E$};
            \node (N_Y) [latent, below=of Y] {$N_Y$};
            \node (N_E) [latent, below=of X_E] {$N_E$};
            \edge[] {Y, N_E} {X_E};
            \edge {N_Y} {Y};
        \end{tikzpicture}
        \caption{Anticausal learning}
        \label{fig:anticausal_learning}
    \end{subfigure}
    \caption{Assuming ICM, in a causal learning setting (a) SSL should be impossible as $P(X_C)$ contains no information about $P(Y|X_C)$, whereas in an anticausal learning setting (b) $P(X_E)$ may contain information about $P(Y|X_E)$ and SSL is thus, in principle, possible.}
    \label{fig:background}
    \vspace{\negspacefig}
\end{figure}

For the task of predicting a target variable $Y$, \citet{Schoelkopf2012} distinguish between \emph{causal learning} (Fig.~\ref{fig:causal_learning}) where all features $X_C$ are causes of $Y$, i.e.,
\begin{align*}
X_C&:=N_C, \\ 
Y&:=f_Y(X_C,N_Y),
\end{align*}
and \emph{anticausal learning} (Fig.~\ref{fig:anticausal_learning}) where all features $X_E$ are effects of $Y$, i.e.,
\begin{align*}
Y&:=N_Y,\\
X_E&:=f_E(Y,N_E).
\end{align*}
In a causal learning setting, it then follows from the ICM principle that $P(X_C)$ and $P(Y|X_C)$ are algorithmically independent, i.e., share no information.
Recalling the goal of improving $P(Y|X)$ from $P(X)$, SSL should thus be impossible.
In the anticausal direction, on the other hand, this algorithmic independence relation is between $P(Y)$ and $P(X_E|Y)$.
Hence, $P(X_E)$ may (and in some cases provably will, see \cite{daniuvsis2010inferring}) share information with $P(Y|X_E)$, and SSL is thus, in principle, possible.

\section{SSL WITH CAUSE AND EFFECTS FEATURES} \label{sec:method}
\begin{figure}[]
    \centering
    \begin{tikzpicture}
        \centering
        \node (X_C) [obs] {$X_C$};
        \node (Y) [obs, right=of X_C] {$Y$};
        \node (X_E) [obs, right=of Y] {$X_E$};
        \node (N_C) [latent, below=of X_C] {$N_C$};
        \node (N_Y) [latent, below=of Y] {$N_Y$};
        \node (N_E) [latent, below=of X_E] {$N_E$};
        \node (f_Y) [const, right=of X_C, yshift=-1.5em,xshift=-1em] {$f_Y$};
        \node (f_E) [const, right=of Y, yshift=-1.5em,xshift=-1em] {$f_E$};
        \edge {N_C} {X_C};
        \edge[] {X_C, N_Y} {Y};
        \edge[] {Y, N_E} {X_E};
        \path[->] (X_C) edge[bend right=-40] node[yshift=1em] {} (X_E);
    \end{tikzpicture}
    \caption{In the SSL setting considered in this work, features can be partitioned into two disjoint sets: potential causes $X_C$ and potential effects $X_E$ of the target $Y$.}
    \label{fig:causeeffect}
    \vspace{\negspacefig}
\end{figure}
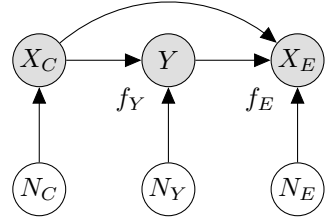

\vspace{\negspace}
In this work, we consider a semi-supervised learning setting where both causes and effects of the target $Y$ are available as features. 
Specifically, we assume that we are given a small labelled sample $(\mathbf{X}^l_C, \mathbf{y}^l, \mathbf{X}^l_E)=\{(\mathbf{x}_c^i,y^i,\mathbf{x}_e^i)\}_{i=1}^{n_l}$ and a large unlabelled sample $(\mathbf{X}^u_C, \mathbf{X}^u_E)=\{(\mathbf{x}_c^i,\mathbf{x}_e^i)\}_{i=n_l+1}^{n_l+n_u}$ generated from an SCM of the form:
\begin{align}
X_C&:=N_C,\label{eq:scmC}  \\ 
Y&:=f_Y(X_C,N_Y),\label{eq:scmY}\\
X_E&:=f_E(X_C,Y,N_E).\label{eq:scmE}
\end{align} 
This causal model is shown in Fig.~\ref{fig:causeeffect}.
We refer to $X_C$ as causal features and $X_E$ as effect features and assume this partitioning to be known a priori (e.g., think of the medical example with risk factors and diagnostic tests).

This setting generalises the cases of causal and anticausal learning considered by \citet{Schoelkopf2012} without positing any new statistical (conditional) independencies---recall that assumptions lie in the \textit{missing} arrows, not in the \textit{present} ones.\footnote{For example, omitting the link $X_C\rightarrow X_E$ renders the two feature sets conditionally independent given $Y$ \citep{vonkugelgen2019semi}, which is restrictive for realistic scenarios, and can already be well addressed by approaches like co-training \citep{Blum1998} which are tailored to such assumptions.}
In particular, the set of causal features $X_C$ may also contain  ``spouse features'' $X_S\subseteq X_C$ which do not directly influence the target, $X_S\not\rightarrow Y$, but only its effects, $X_S\rightarrow X_E$.
All features in the Markov blanket of $Y$ can thus be expressed as members of either $X_C$ or $X_E$, so that our setting remains widely applicable.

Analogous to \eqref{eq:causalmarkov}, the SCM \eqref{eq:scmC}--\eqref{eq:scmE} induces an observational distribution which factorises over the causal graph:
\begin{equation}\label{eq:observationaljoint}
    P(X_C, Y, X_E) = P(X_C)P(Y|X_C)P(X_E|Y,X_C).
\end{equation}
Following \citet{Schoelkopf2012}, we assume the principle of ICM, i.e., that the factors on the RHS of \eqref{eq:observationaljoint} constitute algorithmically-independent causal mechanisms.

Our goal is to predict the target $Y$ from features $(X_C,X_E)$, so we are interested in estimating
\begin{equation}\label{eq:conditional}
    p(y|x_C,x_E)
    =\frac{p(y|x_C)p(x_E|y,x_C)}{\sum_{y' \in \mathcal{Y}}p(y'|x_C)p(x_E|y',x_C)}
\end{equation}
while having additional information about $P(X_C,X_E)$ from unlabelled data.

In analogy to the case of causal learning (see \cref{sec:causalanticausal}), by the ICM principle, the distribution over causes $P(X_C)$ does not contain any information about $P(Y|X_C)$ or $P(X_E|Y,X_C)$ (see RHS of \eqref{eq:observationaljoint}), and thereby also not about $P(Y|X_C,X_E)$ (see ~RHS of \eqref{eq:conditional}).
Indeed, $P(Y|X_C,X_E)$ is completely determined by the structural equations \eqref{eq:scmY} for $Y$ as a function of $X_C$, and \eqref{eq:scmE} for $X_E$ as a function of $X_C$ and $Y$, and does not depend on \eqref{eq:scmC}, i.e., what distribution of causal features $X_C$ is fed into this generative process.\footnote{
This suggests the possibility to perform SSL with covariate-shifted causal features, see, e.g., \citet{vonkugelgen2019semi}.}

Having established that $P(X_C)$ does not contain useful information for our task,\footnote{Since we generally aim to minimise a risk, or \emph{expected} loss, $P(X_C)$ can still help get a better estimate of the expectation operator \citep{Peters2017}. By useful information here we mean information about $P(Y|X_C,X_E)$, see \cref{sec:discussion} for further discussion.}
we are left with $P(Y,X_E|X_C)$ which according to the chain rule of probability admits two possible factorisations,
\begin{align}
    P(Y,X_E|X_C)&=P(Y|X_C)P(X_E|X_C,Y)\label{eq:causalfac}\\
    P(Y,X_E|X_C)&=P(X_E|X_C)P(Y|X_C,X_E). \label{eq:noncausalfac}
\end{align}
Equation \eqref{eq:causalfac} is a \textit{causal} factorisation into independent mechanisms which do not share any information.
Equation \eqref{eq:noncausalfac}, however, corresponds to a {\em non-causal} factorization, implying that the factors on the RHS may share information.
Since we care about estimating $P(Y|X_C,X_E)$ and we have additional information about $P(X_E|X_C)$ from unlabelled data, it is precisely this potential dependence or link between $P(Y|X_C,X_E)$ and $P(X_E|X_C)$ that SSL approaches should aim to exploit in our setting (Fig.~\ref{fig:causeeffect}). We formulate this result as follows.

\paragraph{Main insight.} When learning with both causes and effects of a target as captured by the causal model in~\eqref{eq:scmC}--\eqref{eq:scmE}, $P(X_E|X_C)$ contains all relevant information provided by additional unlabelled data $(\mathbf{X}_C^u,\mathbf{X}_E^u)$ about $P(Y|X_C,X_E)$.
Therefore, SSL approaches for such a setting should aim at exploiting this information by linking these two conditional distributions via suitable additional assumptions.\footnote{Our statement is about how SSL should proceed on the level of distributions, i.e., in the general case; to develop specific SSL algorithms, more concrete assumptions are necessary.}

We remark that this contains previous results for SSL in causal and anticausal learning settings as special cases: in absence of causal features (i.e., for anticausal learning) $P(X_E|X_C)$ reduces to the known setting of $P(X_E)$ containing information about $P(Y|X_E)$, whereas in absence of effect features (i.e., for causal learning) $P(X_E|X_C)$ becomes meaningless, and SSL thereby impossible, both consistent with the findings of \citet{Schoelkopf2012}.

However, our result goes further than this: having additional unlabelled data of both cause and effect features can be strictly more informative than having only unlabelled effects.
To illustrate this point, consider the following thought experiment.
Suppose that $X_E$ is a noisy copy or proxy label of $Y$, i.e., $X_E:=Y + N_E$.
In this case, unlabelled data contains information which is very similar to the information contained in the labelled data.
Learning to predict $X_E$ from $X_C$ requires predicting $Y$ from $X_C$ and can thus be very helpful to solve the problem.

\vspace{\negspace}
\section{NEW ASSUMPTIONS FOR SSL}
\label{sec:assumptions}
\vspace{\negspace}
We now use our insight to reformulate, or refine, standard SSL assumptions (see \cref{sec:background_ssl}) for the setting of Fig.~\ref{fig:causeeffect} where both $X_C$ and $X_E$ are observed.
Our aim is to adapt these assumptions such that they make use of potential information shared between $P(Y|X_C,X_E)$ and the conditional $P(X_E|X_C)$---as opposed to the marginal $P(X_C,X_E)$.

While the previous analysis (\cref{sec:method}) applies to general prediction tasks including regression, we now focus on classification.
For conceptual simplicity and ease of illustration, we will assume binary classification in what follows, but extensions to the multi-class setting are straightforward.
For binary $Y\in\{0,1\}$, we can rewrite \eqref{eq:scmY} and \eqref{eq:scmE} as:
\begin{align}
Y&:=\mathbb{I}\{g(X_C)>U\}\label{eq:scm_classY}\\
X_E&:=  Y f_1(X_C,N_E) + (1-Y) f_0(X_C,N_E) \label{eq:scm_classE}
\end{align}
where $\mathbb{I}$ is the indicator function and $U$ is a uniform random variable on $[0,1]$, so that $g(X_C)$ computes $P(Y=1|X_C)$. 
Allowing arbitrary $g, f_0, f_1$ and $N_E$, this comes without loss of generality.

\subsection{CONDITIONAL CLUSTER ASSUMPTION}
\label{sec:conditional_cluster_assumption}
\vspace{\negspacesub}
While the standard cluster assumption advocates for sharing labels within clusters in the marginal distribution of all features, in view of the above we  postulate that \emph{points in the same cluster of $X_E|X_C$ share the same label $Y$}.
We refer to this as the \emph{conditional cluster assumption}.

One can think of clusters of $X_E|X_C$ by considering the space $\mathcal{F}$ of functions $f:\mathcal{X}_C\rightarrow\mathcal{X}_E$ computing effects from causes.
Different functions in this space correspond, e.g., to different choices of $Y$ and $N_E$ in \eqref{eq:scm_classE}.
The conditional cluster assumption can then be understood as saying that the class-dependent mechanisms $f_0$ and $f_1$ correspond to cluster centroids in $\mathcal{F}$.\footnote{Note that due to the general form of \eqref{eq:scm_classE}, it is possible to have more than one cluster per class in $\mathcal{F}$. For handwritten digits, for example, $N_E$ could act as a switch between 7s with and without the horizontal stroke.}

\begin{figure}
   \centering
    \includegraphics[width=\columnwidth]{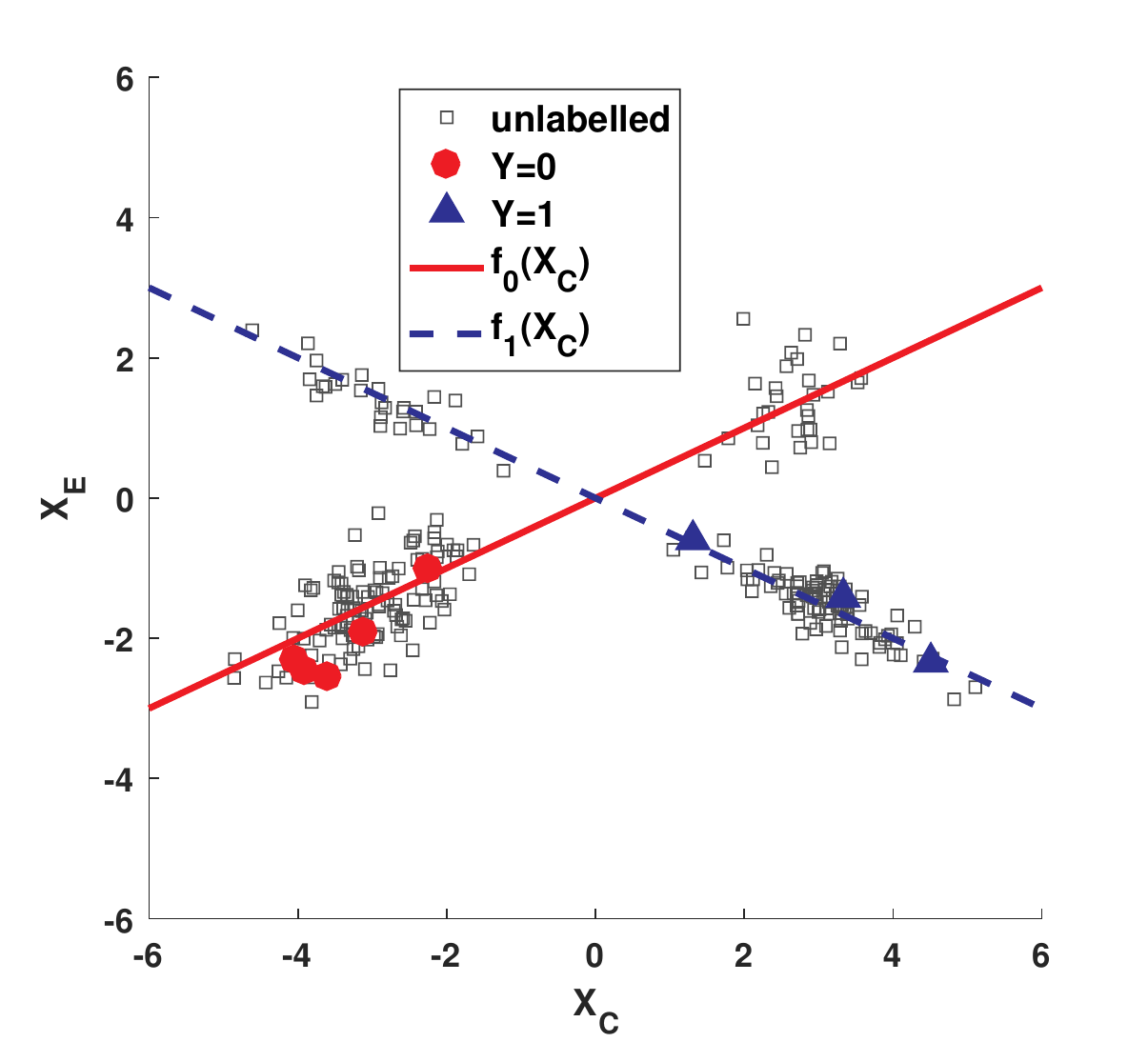}
    \caption{An example dataset arising from our setting with linear $f_0$ and $f_1$ in \eqref{eq:scm_classE}. The conditional cluster assumption links class labels to membership in clusters of $X_E|X_C$, suggesting to classify unlabelled points $(\mathbf{x}_C,\mathbf{x}_E)$ according to whether $\mathbf{x}_E$ is better explained by $f_0(\mathbf{x}_C)$ (solid red line) or $f_1(\mathbf{x}_C)$ (dashed blue line). Conventional SSL methods can easily fail in this case: label propagation or maximum margin would both classify by $\text{sgn}(x_C)$ yielding $\approx$50\% error.}
    \label{fig:linex}
    \vspace{\negspacefig}
\end{figure}

This idea is illustrated in Fig.~\ref{fig:linex} for linear $f_0$ and $f_1$.
In this simple example, knowing that $X_E$ causally depends on $X_C$ (combined with assumptions about the type of dependence, see \cref{sec:conditional_self_learning}) can help identify the true mechanisms (solid red and dashed blue lines), and therefore the correct labelling.
Standard SSL approaches (see \cref{sec:background_ssl}) agnostic to the causal structure, on the other hand, can easily fail in this situation:
large-margin methods (TSVMs) learn to classify $Y$ by the sign of $X_C$ (i.e., using the boundary $X_C=0$), while (unconditional) clustering or graph-based approaches yield a similar classification due to measuring similarity in the joint feature space $(X_C,X_E)$, rather than in terms of the conditional $X_E|X_C$;\footnote{E.g., the points ($-$4,$-$2) and (4,2) are far away in terms of marginal distance, but close in terms of the conditional as measured, e.g., by $||x_E - f_0(x_C)||$ with $f_0(x)=0.5x$, see~\cref{sec:conditional_self_learning}.}
in either case this leads to an error rate of almost 50\%.

\begin{figure}[tb]
   \centering
    \includegraphics[width=\columnwidth]{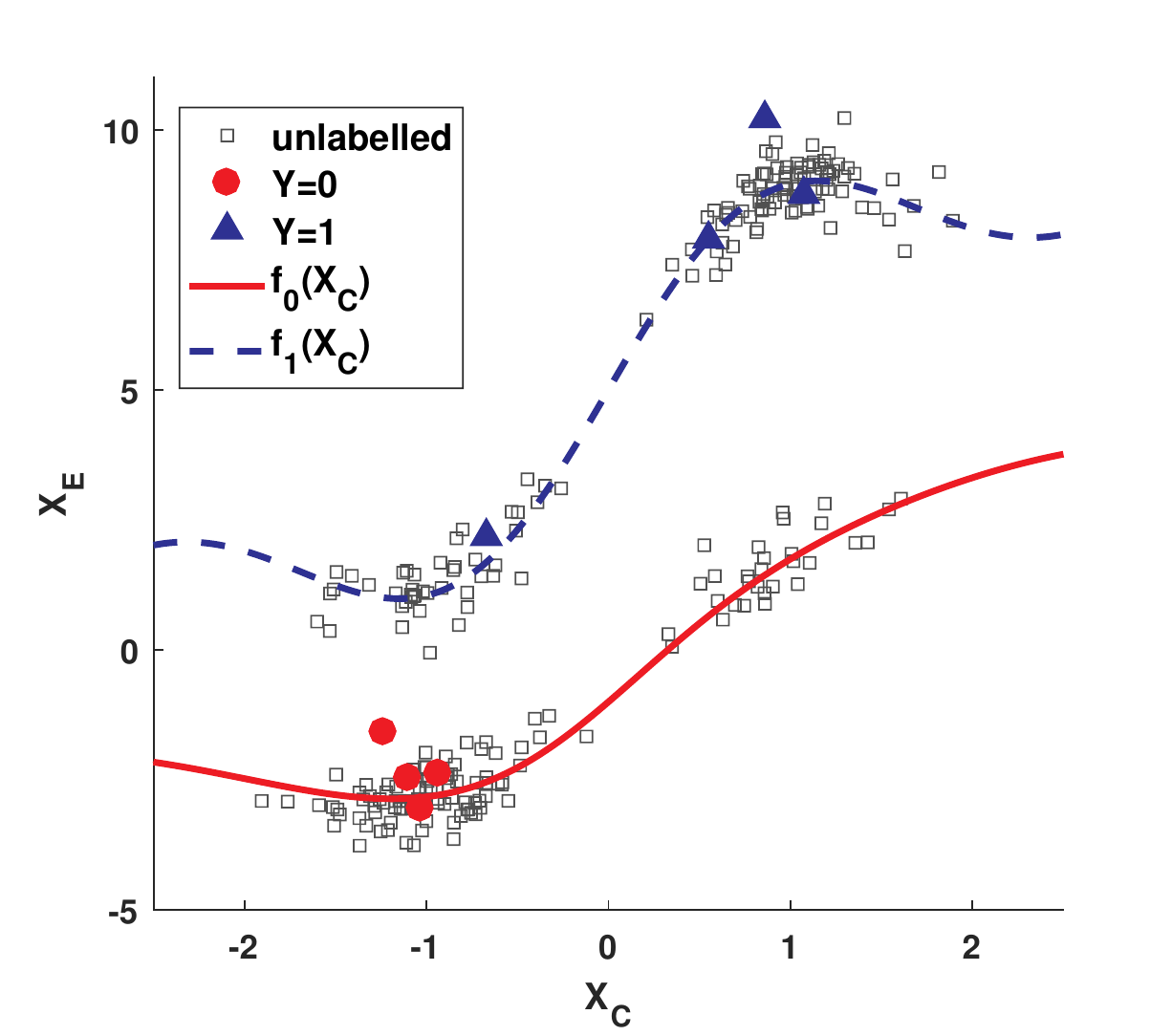}
    \caption{An example with non-linear mechanisms $f_0$ and $f_1$ in \eqref{eq:scm_classE}. Under additional assumptions such as additive unimodal noise, the causal structure induces an asymmetry between features which imposes constraints on the cluster assignments. E.g., the two clusters around $X_C=1$ cannot both be explained by $f_1$, though this would be possible with the roles of $X_C$ and $X_E$ interchanged (i.e., if the figure were rotated by 90$^\circ$).}
    \label{fig:nonlinex}
    \vspace{\negspacefig}
\end{figure}

More generally, knowing the causal partitioning of features into $X_C$ and $X_E$ introduces an asymmetry between features that is not present in standard SSL methods.
In particular, the fact that functions are not allowed to be \textit{one-to-many} provides a constraint.
This becomes apparent for non-linear mechanisms $f_0$, $f_1$ as shown in Fig.~\ref{fig:nonlinex}.
There, the roles of $X_C$ and $X_E$ cannot be exchanged (i.e., the figure rotated by 90$^\circ$) since $f_0$ and $f_1$ are not invertible, thus restricting the possibilities of clustering the data. 

\subsection{LOW-CONDITIONAL-DENSITY SEPARATION ASSUMPTION}
\vspace{\negspacesub}
In a similar vein to \cref{sec:conditional_cluster_assumption}, we can also adapt the low-density separation assumption to our setting.
While in its original form, low-density separation is a statement about the joint density of all features, we have argued that (subject to the ICM principle) $P(X_C)$ contains no information about $P(Y|X_C,X_E)$, but that the conditional $P(X_E|X_C)$ may do so.
We therefore propose that a more justified notion of separation is that \emph{class boundaries of $P(Y|X_C,X_E)$ should lie in regions where $P(X_E|X_C)$ is small}.
We refer to this is as \emph{low-conditional-density separation}.

\vspace{\negspace}
\section{ALGORITHMS}
\label{sec:algorithms}
\vspace{\negspace}
While the main contribution of this work is conceptual, it is illustrative to discuss the implication of our assumptions from \cref{sec:assumptions} for some of the standard approaches to SSL introduced in \cref{sec:background_ssl} and propose variations thereof which explicitly aim to make use of the information shared between the two conditionals $P(X_E|X_C)$ and $P(Y|X_C,X_E)$.

\subsection{SEMI-GENERATIVE MODELS}
\label{sec:semigenerative}
\vspace{\negspacesub}
First, we reconsider the generative approach to SSL.
While a na\"ive model would unnecessarily fit the full distribution---including the uninformative part $P(X_C)$---this approach to SSL is easily adapted to our new assumptions by only modelling the informative part of the generative process, $P(Y,X_E|X_C)$.
This type of \emph{semi-generative model} has been introduced by \citet{vonkugelgen2019semi} in the context of domain adaptation, and under the restrictive assumption of conditionally independent feature sets, $X_C\indep X_E|Y$. 
Here, we relax this assumption, and propose a new approach to fit the resulting model.

Given a semi-generative model parameterised by $\bm{\theta}$, a maximum likelihood approach similar to \eqref{eq:generative_modelSSL} then yields
\begin{equation*}\label{eq:semigenerative}
\textstyle
\argmax_{\bm{\theta}} p(\mathbf{y}^l,\mathbf{X}_E^l|\mathbf{X}_C^l;\bm{\theta}) \sum_{\mathbf{y}^u}p(\mathbf{y}^u,\mathbf{X}_E^u|\mathbf{X}_C^u;\bm{\theta}).
\end{equation*}
Equivalently, we minimise the negative log-likelihood (NLL) which \emph{for fixed labels} decomposes according to~\eqref{eq:observationaljoint} into two separate terms which can be optimised independently for $\bm{\theta}_Y$ and $\bm{\theta}_E$:
\begin{align*}\label{eq:loglik}
    \textstyle
    &\text{NLL}(\bm{\theta}|\mathbf{X}_C,\mathbf{y},\mathbf{X}_E) := -\log p(\mathbf{y},\mathbf{X}_E|\mathbf{X}_C;\bm{\theta})\\
    =& -\log p(\mathbf{y}|\mathbf{X}_C;\bm{\theta}_Y)-\log p(\mathbf{X}_E|\mathbf{X}_C,\mathbf{y};\bm{\theta}_E)
\end{align*}
This separation leads us to an EM-like approach \citep{dempster1977maximum} to find a local optimum of the NLL by iteratively computing the expected label given the current parameters (E-step), and then minimising the NLL w.r.t.\ to the parameters keeping the labels fixed (M-step).
For the setting of hard labels (i.e., $y\in\{0,1\}$), this procedure is summarised in Algorithm \ref{alg:pseudoEM} where for brevity we omit explicitly conditioning the NLL on $\mathbf{X}_C$ and $\mathbf{X}_E$, and where \textit{converged} means $\bm{y}^{(t)}=\bm{y}^{(t-1)}$.
For the specific case of logistic regression for $P(Y|X_C)$ and a class-dependent linear Gaussian model for $P(X_E|X_C,Y)$ we provide a more detailed procedure for both soft and hard labels in Algorithm \ref{alg:ridgeEM} in Appendix~\ref{app:algorithms}.

\begin{algorithm}[t]
\SetAlgoLined
\KwIn{labelled data $(\mathbf{X}^l_C, \mathbf{y}^l, \mathbf{X}^l_E)$; unlabelled data $(\mathbf{X}^u_C, \mathbf{X}^u_E)$; parametric models $p(y|\mathbf{x}_C;\bm{\theta}_Y)$ and $p(\mathbf{x}_E|\mathbf{x}_C,y;\bm{\theta}_E)$}
\KwOut{fitted labels $\mathbf{y}^u$; estimates $\hat{\bm{\theta}}_Y, \hat{\bm{\theta}}_E$}
\BlankLine
$t \leftarrow 0$\\
$\hat{\bm{\theta}}_Y^{(0)}\leftarrow \argmin \text{NLL}(\bm{\theta}_Y|\mathbf{y}^l)$\\
$\hat{\bm{\theta}}_E^{(0)}\leftarrow \argmin \text{NLL}(\bm{\theta}_E|\mathbf{y}^l)$\\
\While{not converged}{
  $\mathbf{y}^{(t)} \leftarrow \mathbb{I}\{p(\mathbf{y}|\mathbf{X}^u_C, \mathbf{X}^u_E; \bm{\theta}_Y^{(t)}, \bm{\theta}_E^{(t)}) > 0.5\}$\\
  $\hat{\bm{\theta}}_Y^{(t+1)}\leftarrow \argmin \text{NLL}(\bm{\theta}_Y|\mathbf{y}^l,\mathbf{y}^{(t)})$\\
  $\hat{\bm{\theta}}_E^{(t+1)}\leftarrow \argmin \text{NLL}(\bm{\theta}_E|\mathbf{y}^l,\mathbf{y}^{(t)})$\\
    $t \leftarrow t+1$\\
 }
 \Return{$\mathbf{y}^{(t-1)}$, $\bm{\theta}_Y^{(t)}$, $\bm{\theta}_E^{(t)}$ }
 \caption{EM-like algorithm for fitting a semi-generative model by maximum likelihood}
 \label{alg:pseudoEM}
\end{algorithm}

\subsection{CONDITIONAL SELF-LEARNING}
\label{sec:conditional_self_learning}
\vspace{\negspacesub}
The second algorithm we propose aims to use the conditional cluster assumption without postulating a generative model in parametric form.
It is loosely related to the ideas of label propagation and self-learning \citep{Scudder65, zhu2002learning}.
However, instead of propagating labels based on similarities between points computed in the joint feature space $(X_C,X_E)$ as in the conventional approach, we instead focus on extracting information contained in $P(X_E|X_C)$.
To this end, we need to place restrictions on the class of functions $f_0$, $f_1$ allowed in \eqref{eq:scm_classE}.
In the following, we therefore assume an \emph{additive noise model} \citep{HoyJanMooPetetal09} defined by
\begin{equation}\label{eq:ANM}
    f_i(X_C,N_E) = f_i(X_C) + N_{E,i} \quad \text{for} \quad i=0,1.
\end{equation}
Fig.~\ref{fig:nonlinex} shows an example of data generated in this way.\footnote{Other choices are, of course, possible: another interesting and more flexible option is the post-nonlinear model of \citet{zhang2009identifiability}, $g_i(f_i(X_C)+N_{E,i})$.}
Note that, unlike in the probabilistic approach of \cref{sec:semigenerative}, we do not make additional assumptions about the exact noise distribution, such as Gaussianity.
We do, however, assume that the noise has zero mean and is unimodal, so that there is one function from $X_C$ to $X_E$ for each label.

Our approach then aims at learning these functions, and can be summarised as follows.
First, we initialise $\hat{f}_0$ and $\hat{f}_1$ from labelled data by regressing $X_E$ on $X_C$.
Next, we compute the predictions of the $\hat{f}_i$ on the unlabelled sample, label the point with smallest prediction error as the respective class, and use it to update the corresponding $\hat{f}_i$.
This procedure is repeated until all initially unlabelled points are labelled.
We refer to this approach as \emph{conditional self-learning}, and summarise it in Algorithm \ref{alg:conditionalprop}. The notation $\mathbf{X}_{C,i}$ and $\mathbf{X}_{E,i}$ refers to those samples of $X_C$ and $X_E$ with label $Y=i$, $\mathbf{r}_i$ in step 5 denotes the vector of residuals from regressing $\mathbf{X}_E^u$ on $\mathbf{X}_C^u$ using $\hat{f}_i$, and $\mathbf{r}_{i,j}$ in step 7 denotes the $j^{th}$ such residual.

\begin{algorithm}[t]
\SetAlgoLined
\KwIn{labelled data $(\mathbf{X}^l_C, \mathbf{y}^l, \mathbf{X}^l_E)$; unlabelled data; $(\mathbf{X}^u_C, \mathbf{X}^u_E)$; \texttt{regress}() method}
\KwOut{fitted labels $\mathbf{y}^u$; functions $\hat{f}_0, \hat{f}_1$}
\BlankLine
$t \leftarrow 0$\\
\While{unlabelled data left}{
  \For{$i=0,1$}
  {
  $\hat{f}_i^{(t)}\leftarrow$ \texttt{regress}$(\mathbf{X}^l_{E,i}, \mathbf{X}^l_{C,i})$\\
  $\mathbf{r}_i\leftarrow ||\mathbf{X}_E^u-\hat{f}_i^{(t)}(\mathbf{X}_C^u)||^2$\\
  }
  $(i,j)\leftarrow \argmin \{\mathbf{r}_{i,j}: i=0,1; j=1,...,n_u\}$\\
  $y^{n_l+j}\leftarrow i$\\
  $\mathbf{X}^l_{E,i}, \mathbf{X}^l_{C,i}\leftarrow \texttt{append}(\mathbf{x}_E^{n_l+j},\mathbf{x}_C^{n_l+j})$\\
    $t \leftarrow t+1$\\
    
 }
 \Return{$\mathbf{y}^u,\hat{f}_0^{(t-1)}, \hat{f}_1^{(t-1)}$}
 \caption{Conditional self-learning}
 \label{alg:conditionalprop}
\end{algorithm}

\begin{table*}[t]
\centering
    \ra{1.2}
    \caption{Average accuracies on unlabelled data (higher is better) $\pm$ one standard deviation across 100 runs, each time randomly drawing 10 (for S1, S2, S3) or 20 (for Pima Diabetes and Heart Disease) new labelled and 200 new unlabelled samples. Results refer to transductive evaluation for ease of comparison with other methods.
    The three best methods for each dataset are highlighted in bold. The last four rows are our causally-motivated methods. The ``-'' indicates that label propagation did not converge on S3, and (sup.) indicates purely supervised baselines. 
    \vspace{0pt}}
    
    \label{tab:results}
    \begin{tabularx}{\textwidth}{@{}lXXXXX@{}}
        \toprule
        \textbf{Method} & \textbf{S1 (linear)} & \textbf{S2 (non-linear)} & \textbf{S3 (multi-dim.)} & \textbf{Pima Diabetes} & \textbf{Heart Disease} \\
        \midrule
        Lin.~log.~reg. (sup.) & .968  $\pm$  .023 & .823  $\pm$  .080 & .945  $\pm$  .039 & .626  $\pm$  .058 & .526  $\pm$  .066\\
        Lin.~T-SVM  & .865  $\pm$  .093 & .878 $\pm$  .074 & .822  $\pm$  .117 & .602  $\pm$  .065 & \textbf{.746 $\pm$  .060}\\
        RBF T-SVM   & .863  $\pm$  .094 & .876  $\pm$  .075 & .821 $\pm$ .116 & .601 $\pm$  .064 & \textbf{.745  $\pm$  .060} \\
        RBF label propag. & .924  $\pm$  .082 & .909 $\pm$ .065 & - & .650  $\pm$  .030 & .528  $\pm$  .068\\
        \midrule
        Semi-gen.~(sup.) & .968  $\pm$  .076 & \textbf{.935  $\pm$  .074} & .949 $\pm$ .082 & \textbf{.669  $\pm$  .064} & .550  $\pm$  .096 \\
        Semi-gen.+soft EM & \textbf{.986  $\pm$  .081} & \textbf{.989  $\pm$  .024} & \textbf{.991 $\pm$ .067} & \textbf{.661  $\pm$  .063} & .518  $\pm$  .050 \\
        Semi-gen.+hard EM & \textbf{.985  $\pm$  .079} & \textbf{.972  $\pm$  .058} & \textbf{.987 $\pm$ .076} & \textbf{.695  $\pm$  .064} & .518  $\pm$  .050\\
        Cond. self-learning & \textbf{.980  $\pm$  .052} & .923  $\pm$  .090 &\textbf{.961 $\pm$ .069} & .659  $\pm$  .079 & \textbf{.719  $\pm$  .076}\\
       \bottomrule
    \end{tabularx}
    \vspace{\negspacefig}
\end{table*}

\vspace{\negspacepar}
\paragraph{Connection to the probabilistic approach}
It is also possible to use the above approach with soft labels (as often done in conventional label propagation \citep{zhu2003semi, zhou2004learning}) by using a weighted regression scheme. 
This requires a method of computing regression weights from prediction errors of $\hat{f}_0$ and $\hat{f}_1$, though, and therefore needs additional assumptions or heuristics.
We note that choosing a particular distribution for $N_E$ and using $P(Y|X_C)$ as a class prior leads to a soft-label EM approach similar to \cref{sec:semigenerative}, see Algorithm \ref{alg:ridgeEM} in Appendix \ref{app:algorithms}. We therefore presently restrict ourselves to hard labels.

\vspace{\negspacepar}
\paragraph{Connection to competition of experts}
While it is conceptually based on the ICM assumption and an analysis of the causal structure among the feature set, the conditional self-learning approach is linked to a number of known methods, including not only self-learning, but also methods building on {\em competition of experts}, as recently applied to the problem of learning causal mechanisms. In this work, the functions $f_0, f_1$ are generative models competing for data that has undergone unknown transformations, and eventually each specialising on how to invert one of those transformations \citep{parascandolo2018learning}.

\vspace{\negspacesub}
\section{EXPERIMENTAL RESULTS}
\label{sec:experiments}
\vspace{\negspace}
To corroborate our analysis with empirical evidence, 
we evaluate our algorithms from \cref{sec:algorithms} on synthetic data as well as on two medical datasets from the UCI repository.
We compare with T-SVMs \citep{vapnik1998, joachims1999transductive} with linear and RBF kernels using the \texttt{q3svm} implementation \citep{q3svm}; and with label propagation \citep{zhu2003semi, zhou2004learning} using the implementation in \texttt{scikit-learn} \citep{scikit-learn}.
We use the default hyper-parameters in all cases.
For our conditional self-learning algorithm, we use linear ridge regression with default regularisation strength~1, and for the EM algorithms we use logistic regression for $Y|X_C$ and linear, class-dependent Gaussians for $X_E|X_C,Y$, see Algorithm \ref{alg:ridgeEM} and Appendix~\ref{app:syntheticdata} for details.  

\vspace{\negspacepar}
\paragraph{Synthetic data}
As controlled environments, we generate various synthetic datasets with cause and effect features of three different types (S1, S2, S3): 
S1 represents linearly-separable data, S2 corresponds to datasets with a non-linear decision boundary of which the data shown in Figure \ref{fig:nonlinex} is an example, and S3 is a version of S2 with multi-dimensional features.
Details of how exactly synthetic data is generated are provided in Appendix~\ref{app:syntheticdata}.

\vspace{\negspacepar}
\paragraph{Medical data}
As real-world data, due to the fact that both plausibly contain cause and effect features, we choose the two medical datasets \emph{Pima Indians Diabetes} \citep{smith1988using} and \emph{Heart Disease} \citep{detrano1989international}.
We select those features which are most strongly correlated with the target variable ($p<0.01$), and categorise them into cause and effect features to the best of our knowledge, see Appendix~\ref{sec:medical_data} for details.

\vspace{\negspacepar}
\paragraph{Results}
The results of our experiments are summarised in Table \ref{tab:results}, see the table caption for simulation details.
On the synthetic datasets, our causally-motivated methods outperform the purely supervised logistic regression baseline as well as the other SSL approaches, which in the case of S1 and S3 even lead to decreased accuracy.
The probabilistic approaches perform particularly well on the synthetic datasets, which was expected since the generative model for these cases was specified by us, and its correct form thus known.
Conversely, on the Heart Disease dataset they degraded performance, presumably due to model misspecification.

Conditional self-learning, based on the weaker assumption of additive noise, performs competitively, in particular on the real data. 
Notably, it is the only method that improves upon the supervised logistic-regression baseline (i.e., achieves SSL) for all five datasets considered.

As a sanity check, we also ran our causally inspired methods with the roles of cause and effect features exchanged which led to deteriorated performance (results not shown). 
This observation provides additional support for our analysis that algorithmic information is shared only between the conditionals $P(Y|X_C,X_E)$ and $P(X_E|X_C)$.

\vspace{\negspacesub}
\section{DISCUSSION}
\label{sec:discussion}
\vspace{\negspace}
The present paper looks at SSL from the perspective of causal modelling.
We argue that if we know that the input $X$ can be partitioned into cause and effect features ($X_C$ and $X_E$), then this has surprising theoretical implications for how SSL should utilise unlabelled data:
rather than simply exploiting links between the marginal $P(X)$ and the conditional $P(Y|X)$ (as formalised, e.g., in the standard SSL cluster assumption), one should exploit links between two conditional distributions, $P(X_E|X_C)$ and $P(Y|X_C,X_E)$.
In other words, SSL should not blindly be applied to the joint feature set, but instead only to effect (or confounded, see below) features while conditioning on purely causal ones.\footnote{Interestingly, such conditioning may allow to extend SSL to settings where the missingness mechanism depends on $X_C$, i.e., covariate-shifted or missing-at-random (MAR) data \citep{Sugiyama2012, mohan2013graphical}.}
Note that we view this not as a contradiction to the usual cluster assumption, but rather as an explication or refinement thereof, taking into account the causal structure; indeed, it subsumes SSL in the anticausal setting as a special case. It does not subsume SSL in the causal setting, but, as argued by \citet{Schoelkopf2012}, SSL is futile in this case.

\vspace{\negspacepar}
\paragraph{Impossibility of SSL in the causal direction} For certain settings, SSL yields provable improvements over a purely-supervised solution, irrespective of the causal structure,
in terms of the surrogate loss in a transductive setting
\citep{loog2015contrastive, krijthe2017projected, krijthe2018pessimistic}.
For linear discriminant analysis, for instance, such improvements are obtained almost surely.
This may appear at odds with the result that SSL in the causal direction is impossible.
The apparent contradiction can, however, be explained as follows.
The claimed lack of useful information (subject to ICM) contained in $P(X_C)$, as used in the analysis of \citet{Schoelkopf2012} and in the present work, refers to information useful \textit{for improving our estimate of} $P(Y|X)$.
However, the success of SSL is often measured instead in terms of the \textit{risk}, i.e., the expected loss w.r.t.\ the \textit{joint} distribution.
Thus, even though knowing $P(X_C)$ in a causal learning setting does not help make better predictions \textit{for any given} $x$, it may still help to achieve a lower risk.
In a sense, knowing $P(X_C)$ helps decide which parts of the support are most important to get right.
This, in turn, can be particularly helpful when dealing with model misspecification \citep{urner2011access}. 
For further discussion of this point, we refer to \citet[][Section 5.1.2]{Peters2017}.
Moreover, we remark that most settings for which SSL has proven particularly useful empirically fall in the category of anticausal learning problems, e.g, recognising digit identity $Y$ from its handwritten realisation $X_E$ \citep[e.g.,][]{kingma2014semi} or classifying document topics $Y$ from the contained words $X_E$ \citep[e.g.,][]{nigam2006semi}. 

\vspace{\negspacepar}
\paragraph{Validity of ICM and the role of latent confounding}
Our work uses the assumption of independent causal mechanisms (ICM) generating the data. 
While well-motivated from the independence of separate physical processes, in practice, the ICM principle may not hold \textit{exactly}: for example, some mechanisms may have co-evolved to be algorithmically dependent \citep{Peters2017}.
While latent confounding does not \textit{a priori} invalidate our assumption---recall that ICM is not a statement about \textit{statistical} but \textit{algorithmic} independence on the level of mechanisms---a confounder $H$ of $X_C$ and $Y$ can nonetheless introduce such dependence due to its shared influence on the two distributions:
\begin{align*}
  P(X_C)&=\textstyle\int P(X_C|H)\, dP(H),\\
  P(Y|X_C) &= \textstyle\int P(Y|X_C, H)\, dP(H).
\end{align*}
Methods for detecting hidden confounding \citep[e.g.,][]{hoyer2008estimation, janzing2018detecting} can potentially help identify such confounded causal features, which should then also be treated as containing useful information.
In weakly-confounded settings, however, ICM can still be a useful principle, and has proven to be so in applications, e.g., in cause-effect inference.
Moreover, many causal features such as age or sex in our medical example can reasonably be assumed to not be caused by other variables, and can therefore safely be treated as unconfounded.
We also note that since SSL is about prediction in an i.i.d.\ setting, latent confounding is generally less problematic than for classical causal inference tasks, such as predicting effects of interventions.
Considering features which are neither causes nor effects, but are correlated to the target by unobserved confounders, constitutes an interesting extension for future work.

\vspace{\negspacepar}
\paragraph{Known causal partitioning}
Another important assumption for our analysis is knowledge about the causal structure.
While we do not address the separate problem of causal discovery, we argue that often sufficient domain knowledge is available to identify maybe not a fine-grained causal graph, but at least a partitioning of features into disjoint sets of potential causes and potential effects.
The medical setting described in the introduction constitutes such a real-world example.
We also note that under certain assumptions (e.g.,\ additive noise, as for the version of conditional self-learning presented here) causal direction can be identified from purely observational data \citep{HoyJanMooPetetal09}.
Such techniques applied to $\{X_C,X_E\}$ may help reveal the causal nature of features, even with only few labels.

\vspace{\negspacepar}
\paragraph{Connections to domain adaptation}
Since our proposed approach is robust to (domain-induced) changes in $P(X_C)$, provided that $P(Y|X_C)$ and $P(X_E|Y,X_C)$ remain invariant, exploring it further in the context of domain adaptation constitutes an interesting future direction (see also footnotes 7 and 13).
Data from different environments may provide valuable causal clues that can potentially be leveraged by constraint- \citep{huang2019causal, magliacane2018domain}, invariance- \citep{Rojas-Carulla2018}, or variance-penalty--based approaches \citep{heinze2017conditional} to learn the underlying causal structure and identify subsets of features for which $P(Y|X_C)$ and $P(X_E|Y,X_C)$ remain stable.

\vspace{\negspacesub}
\section{CONCLUSION}
\label{sec:conclusion}
\vspace{\negspace}
While the present analysis is intriguing and points out a previously unexplored link between two conditional distributions, the jury is still out on how in general to best exploit unlabelled data in machine learning.
The present insight is but one step, and in particular, while encouraging, the algorithms and experiments based upon it can only be a starting point (see Appendix \ref{app:nonlinear} for proposed extensions).
They may lead to new approaches that make explicit use of causal structure and exploit the conditional cluster assumption (and potentially the low-conditional-density separation assumption) in more elegant and effective ways.
Ultimately, the value of novel assumptions and conceptual models lies in whether they provide a fertile basis to inspire further algorithm development and theoretical understanding. We hope that the present ideas and analysis will constitute such a contribution.

\vspace{\negspacesubsub}
\subsubsection*{Acknowledgements}
\vspace{\negspacesubsub}
The authors thank Atalanti Mastakouri, Jonathan Gordon, Luigi Gresele, Sebastian Weichwald, and the anonymous reviewers for helpful discussions and suggestions.

\clearpage
\printbibliography

\clearpage
\appendix
\section{Algorithms} \label{app:algorithms}
Algorithm \ref{alg:ridgeEM} describes concrete soft and hard labelling versions of the EM approach proposed in Algorithm \ref{alg:pseudoEM} for the model assumption of a logistic regression for $Y|X_C$ and linear Gaussian distributions for $X_E|X_C,Y=0$ and $X_E|X_C, Y=1$ in some more detail. It was used for our experiments in \cref{sec:experiments}.

\section{Experimental details}
\subsection{Synthetic datasets}
\label{app:syntheticdata}

The synthetic datasets used in our experiments were generated as follows.
First, we draw $X_C\in\mathbb{R}^{d_C}$ from a mixture of $m$ $d_C$-dimensional Gaussians.
Next, we draw $Y\in \{0,1\}$ and $X_E\in\mathbb{R}^{d_E}$ according to the SCM
\begin{equation*}
\begin{aligned}\label{eq:synth_data}
Y&:=\mathbb{I}[\sigma(\mathbf{a}^TX_C+b)>N_Y],\\
X_E&:=\begin{cases} 
		 \mathbf{A}_0X_C+\mathbf{b}_0+\mathbf{D}_0N_E \quad \text{if} \quad Y=0,\\
		 \mathbf{A}_1X_C+\mathbf{b}_1 + \mathbf{D}_1N_E \quad \text{if} \quad Y=1,
    	\end{cases}
\end{aligned}
\end{equation*}
with $N_Y\sim U[0,1]$, $N_E\sim\mathcal{N}_{d_E}(0,I)$, $\mathbf{a} \in \mathbb{R}^{d_C}$; $b\in\mathbb{R}$; $\mathbf{A}_0, \mathbf{A}_1 \in \mathbb{R}^{d_E\times d_C}$; and $\mathbf{D}_0, \mathbf{D}_1 \in \mathbb{R}^{d_E\times d_E}$ are diagonal matrices of standard deviations. $\sigma(x)=(1+e^{-x})^{-1}$ denotes the logistic sigmoid function.

This induces the following distributions:
\begin{align*}
    Y|X_C &\sim \text{Bernoulli}(\mathbf{a}^TX_C+b),\\
    X_E|(X_C,Y=i) &\sim \mathcal{N}(\mathbf{A}_iX_C+\mathbf{b}_i, D_i^2)\quad  \text{for} \quad i=0,1.
\end{align*}

For experiments on synthetic data we draw a new dataset according to the above generative process in each run, keeping parameters fixed as follows.

\textit{S1: Linear synthetic dataset
}
\begin{itemize}
    \item feature dimensions: $d_C=d_E=1$
    \item $X_C$: $m=3$ components with weights $w = [0.3, 0.4, 0.3]$, means $\mu_C = [-5, 0, 5]$ and standard deviations $\sigma = [0.5, 0.5, 0.5]$
    \item $Y$: $a=0.5, b=0$
    \item $X_E$: $A_0=A_1=1, b_0=-b_1=2, D_0=D_1=0.25$
\end{itemize}

\textit{S2: Nonlinear synthetic dataset
}
\begin{itemize}
    \item feature dimensions: $d_C=d_E=1$
    \item $X_C$: $m=2$ components with weights $w = [0.5, 0.5]$, means $\mu_C = [-3, 3]$ and variances $\sigma^2 = [0.5, 0.5]$
    \item $Y$: $a=0.5, b=0$
    \item $X_E$: $A_0=-A_1=0.5, b_0=b_1=0, D_0=D_1=0.25$
\end{itemize}

\textit{S3: Nonlinear multidimensional synthetic dataset
}
\begin{itemize}
    \item feature dimensions: $d_C=d_E=2$
    \item $X_C$: $m=2$ components with weights $w = [0.5, 0.5]$, means $[-3, -3], [3,3]$ and covariances $\Sigma=\text{diag}([0.5,0.5])$
    \item $Y$: $a=0.5, b=0$
    \item $X_E$: $a_0=-a_1=0.5, b_0=b_1=0, D_0=D_1=\text{diag}(0.25, 0.25)$
\end{itemize}

\subsection{Medical datasets}
\label{sec:medical_data}
For the Pima Indians Diabetes dataset we use the partitioning  $X_C=$\{DiabetesPedigreeFunction, Pregnancies, BMI\}
and $X_E=$\{Glucose\}. DiabetesPedigreeFunction is a measure of the family history of diabetes and
BMI stands for body mass index.\footnote{The relationship between BMI and Diabetes may actually be cyclic \citep{Feldstein}, though it is believed that the causal link BMI $\rightarrow$ Diabetes is the stronger one.}  

For the (Coronary) Heart Disease dataset we used the partitioning $X_C=$\{sex, ca, thal\} and $X_E=$\{chest pain\}. Here, ``ca'' refers to the number of major vessels (0-3) that contained calcium (colored by flouroscopy), and ``thal'' to thallium scintigraphy results, a nuclear medicine test that images the blood supply to the muscles of the heart.

\section{Non-linear mechanisms}
\label{app:nonlinear}
For simplicity, we have focused on linear mechanisms in our experiments.
More complex choices are, of course, possible, and---depending on the setting---necessary.

Figure \ref{fig:nonlinex} shows an example of a non-linear additive noise model  \citep{HoyJanMooPetetal09} for $f_0$ and $f_1$.
To address this case, one may simply choose a different regression method in our conditional self-learning approach (Algorithm \ref{alg:conditionalprop} in the main paper); kernel-ridge or Gaussian-process regression \citep{williams2006gaussian} are two obvious choices.

For high-dimensional or structured data such as natural images or text, more complex non-additive noise models can be used for $f_i(X_C,N_E)$, such as the conditional versions of GANs or VAEs.
The work of \citet{parascandolo2018learning} mentioned at the end of \cref{sec:algorithms} constitutes an example of the former.


\clearpage
\onecolumn


\begin{algorithm}[t]
\SetAlgoLined
\KwIn{labelled data $(\mathbf{X}^l_C, \mathbf{y}^l, \mathbf{X}^l_E)$; unlabelled data $(\mathbf{X}^u_C, \mathbf{X}^u_E)$; regularisation strength $\lambda$ for ridge regression; labelling type (hard/soft)}
\KwOut{fitted labels $\mathbf{y}^u$; parameters $\bm{\theta}_Y, \bm{\theta}_E$}
\BlankLine

$t \leftarrow 0$\\
\BlankLine
\tcp{initialise parameter estimates using only the labelled sample}
$\bm{\theta}_Y^{(0)}\leftarrow$ \texttt{LogisticRegresssion}$(\mathbf{X}_C^l, \mathbf{y}^l; \bm{\theta}_Y)$\\
$\bm{\theta}_{E,0}^{(0)}\leftarrow ((\mathbf{X}_{C,0}^{l})^T\mathbf{X}_{C,0}^{l}+\lambda I)^{-1}(\mathbf{X}_{C,0}^{l})^T \mathbf{X}_{E,0}^{l}$\\
$\bm{\theta}_{E,1}^{(0)}\leftarrow ((\mathbf{X}_{C,1}^{l})^T\mathbf{X}_{C,1}^{l}+\lambda I)^{-1}(\mathbf{X}_{C,1}^{l})^T \mathbf{X}_{E,1}^{l}$\\
   \BlankLine
\While{not converged}{
     \BlankLine
  \tcp{compute soft (class probabilities) and hard labels (E-step)}
  $\mathbf{q}^{(t)} \leftarrow p(\mathbf{y}|\mathbf{X}^u_C, \mathbf{X}^u_E; \bm{\theta}_Y^{(t)}, \bm{\theta}_{E,0}^{(t)}, \bm{\theta}_{E,1}^{(t)})$\\
  $\mathbf{y}^{(t)} \leftarrow \mathbb{I}\{\mathbf{q}^{(t)}>0.5\}$\\
   \BlankLine
      \BlankLine
   \tcp{compute weights (1 for hard labels, class prob. for soft labels)}
   \eIf{hard labelling}{
  $\mathbf{w}^{(t)}\leftarrow\mathbf{1}$
  }
  {
  $\mathbf{w}^{(t)}\leftarrow[\mathbf{1}; \mathbf{y}^{(t)}\odot\mathbf{q}^{(t)}+ (1-\mathbf{y}^{(t)})\odot(1-\mathbf{q}^{(t)})]$
  }
    $\mathbf{W}_1^{(t)}\leftarrow \text{diag}([\mathbf{y}^l; \mathbf{q}^{(t)}])$\\
  $\mathbf{W}_0^{(t)}\leftarrow \text{diag}([\mathbf{1}-\mathbf{y}^l; \mathbf{1} - \mathbf{q}^{(t)}])$\\
 \BlankLine
    \BlankLine
  \tcp{Update parameter estimates keeping estimated labels fixed (M-step)}
  $\bm{\theta}_Y^{(t)}\leftarrow$ \texttt{WeightedLog.~Reg.~}$([\mathbf{X}_C^l; \mathbf{X}_C^u], [\mathbf{y}^l; \mathbf{y}^{(t)}], \mathbf{w}^{(t)}; \bm{\theta}_Y)$\\
  $\bm{\theta}_{E,1}^{(t)}\leftarrow (\mathbf{X}_{C}^T\mathbf{W}_1^{(t)}\mathbf{X}_{C}+\lambda I)^{-1}\mathbf{X}_{C}^T \mathbf{W}_1^{(t)}\mathbf{X}_{E}$\\
  $\bm{\theta}_{E,0}^{(t)}\leftarrow (\mathbf{X}_{C}^T\mathbf{W}_0^{(t)}\mathbf{X}_{C}+\lambda I)^{-1}\mathbf{X}_{C}^T \mathbf{W}_0^{(t)}\mathbf{X}_{E}$\\
    $t \leftarrow t+1$\\
 }
 \Return{$\mathbf{y}^{(t-1)}$, $\bm{\theta}_Y^{(t)}$, $\bm{\theta}_E^{(t)}$ }
 \caption{Soft and hard label EM algorithms for a semi-generative model with logistic regression for $P(Y|X_C)$ and class-dependent linear Gaussian models for $P(X_E|Y,X_C)$.}
 \label{alg:ridgeEM}
\end{algorithm}

\end{document}